\documentclass{article}

\usepackage[preprint]{neurips_2021}

\usepackage[utf8]{inputenc} 
\usepackage[T1]{fontenc}    
\usepackage{hyperref}       
\usepackage{url}            
\usepackage{booktabs}       
\usepackage{amsfonts}       
\usepackage{nicefrac}       
\usepackage{microtype}      
\usepackage{xcolor}         
\usepackage{amsmath}
\usepackage{graphicx}
\usepackage{caption}
\usepackage{subcaption}
\usepackage{cancel}
\usepackage{placeins}
\usepackage{array}
\usepackage{multirow}

\newcommand\figref{Figure~\ref}
\newtheorem{definition}{Definition}
\newtheorem{theorem}{Theorem}
\newcolumntype{P}[1]{>{\centering\arraybackslash}p{#1}}
\newcolumntype{M}[1]{>{\centering\arraybackslash}m{#1}}
\newcommand{\indep}{\perp \!\!\! \perp}

\title{Self-explaining variational posterior distributions for Gaussian Process models}

\author{%
  Sarem Seitz\\
  Department of Information Systems and Applied Computer Science\\
  University of Bamberg\\
  Bamberg, Germany \\
  \texttt{sarem.seitz@uni-bamberg.de} \\
}

\begin{document}

\maketitle

\begin{abstract}
    Bayesian methods have become a popular way to incorporate prior knowledge and a notion of uncertainty into machine learning models. At the same time, the complexity of modern machine learning makes it challenging to comprehend a model's reasoning process, let alone express specific prior assumptions in a rigorous manner. While primarily interested in the former issue, recent developments in transparent machine learning could also broaden the range of prior information that we can provide to complex Bayesian models. Inspired by the idea of self-explaining models, we introduce a corresponding concept for variational Gaussian Processes. On the one hand, our contribution improves transparency for these types of models. More importantly though, our proposed self-explaining variational posterior distribution allows to incorporate both general prior knowledge about a target function as a whole and prior knowledge about the contribution of individual features.
\end{abstract}

\section{Introduction}
As the field of interpretable and explainable Machine Learning is getting more and more traction, machine learning methods whose reasoning and decision processes were once incomprehensible to human users are finally starting to become transparent. While a general solution to the challenge of humanly tangible, yet sufficiently complex models still seems to be far off in the future, recent developments have yielded promising results. The primary advantages of interpretable models are, as noted in \citet{scienceinterpretable}, (scientific) understanding on the one hand and on the other hand safety, especially operational and ethical safety.

In regards to the former, expressing a complex learning problem in a form that humans can make sense of also presents the chance to let human prior knowledge augment the learning process. Besides the well established field of imitation learning\footnote{see for example \citet{surveyimitationlearning} for an overview}, Bayesian methods are another apparent candidate for such endeavor. The strongest points for the Bayesian route are - first - the ability to express expert knowledge even before any data is available. Second, Bayesian statistics is embedded in a rigorous mathematical foundation that allows to derive theoretical results in a deductive manner. In practical terms, this can be particularly valuable when observational data are sparse or highly expensive to obtain or generate. 

To give a concrete motivational example, consider a simple regression problem where the relation between input and target features is given by, leaving aside potential noise, a linear function. If a reasonably informed expert is aware of this relation and simultaneously able to suitably articulate this prior knowledge to a machine learning model, we can expect the performance of such model to improve over an uninformed counterpart.

While Bayesian methods are commonly praised for their ability to deal with the presence of expert knowledge, the complexity of modern Bayesian models only permits the expression of very general prior beliefs. Consider the case of Gaussian Process (GP) models as arguably the figurehead of Bayesian non-parametrics. The choice of the GP kernel function allows, in theory, to express certain functional prior assumptions. Due to the above mentioned complexity issue however, it is fairly common to use some variation of the squared exponential (SE-) kernel per default. On the one hand, generic prior distributions as implied by the SE-kernel might be the only reasonable choice to describe a complex problem on a global scale. On the other hand though, prior information about more granular properties of the target function is completely discarded under these circumstances.

This brings us back to the initial consideration of leveraging transparent machine learning models in order to mitigate this limitation. For this purpose, we will take advantage of the common approach to have a model learn a globally complex function while being able to locally, for a given instance, decompose a model's decision into the contribution of each feature. Such procedure has been proposed for Neural Networks in particular by \citet{selfexplainnets} who coined the term \textit{self-explaining} models. We will directly transfer their idea to variational approximations for GPs as introduced in \citet{titsiassvgp, hensmanbig} and exploit the resulting structure of the variational posterior. The proposed model is both self-explaining as well and at the same time extends the possibilities to express prior knowledge in the context of GPs.

\section{Transparent Machine Learning}
In regards to transparency in machine learning, terms like \textit{interpretable Machine Learning} or \textit{explainable Artificial Intelligence} (XAI) have become quite widespread and popular. However, actual definitions of such terms still vary from author to author. In our context, we use the following definitions of interpretation and explanation from \citet{understandingdeepnets}:
\begin{definition}
    An \textbf{interpretation} is the mapping of an abstract concept into a domain that the human can make sense of. An \textbf{explanation} is the collection of features of the interpretable domain, that have contributed for a given example to produce a decision.
\end{definition}
Besides images and text being interpretable domains as noted in \citet{understandingdeepnets}, we note that reasonably sized mathematical or statistical models can be considered as being interpretable as well. Take for example the standard linear regression model
\begin{equation}\label{linearmodel}
    y=X\beta  
\end{equation}
and the well-known \textit{interpretation} of each coefficient being the average marginal effect of the corresponding variable or feature. Unless such model contains dozens of relevant variables, a human with sufficient domain knowledge about the problem being modelled can then easily make sense of the resulting qualitative implications. It can also be seen that a linear model provides a globally applicable \textit{explanation} for a given example, i.e. the contribution of each feature is always the same, independently of an example's location in its domain $\mathcal{X}$.

On the other hand, it is obvious that the plain linear model is unable to deal with complex problems in a satisfactory manner, yet problems of high complexity are particularly relevant in machine learning. In order to solve this rather severe shortcoming, a straightforward extension of \eqref{linearmodel} are so-called \textit{varying coefficient models} as first introduced by \citet{hastie1993varying}. Here, the parameters $\beta$ are themselves functions of some covariates $R$:
\begin{equation}\label{varcoeff}
    y=\beta_1(R)^T\cdot X_1 + ... + \beta_K(R)^T\cdot X_K
\end{equation}
where $X_k$ denotes the $k$-th feature column of $X$. For our purpose, we usually have $R\equiv X$. In order to obtain a sufficiently flexible model from \eqref{varcoeff}, using a universal approximator like Neural Networks for the $\beta_k(\cdot)$ is an obvious choice and proposed in particular by \citet{selfexplainnets} under the umbrella term \textit{self-explaining neural networks} (SENN). Although the latter propose an even more general model, the definition in \eqref{varcoeff} as a special case shall suffice for our means. Hence, our model of interest is defined as follows:
\begin{equation}\label{senn}
    y=\phi_1(X)^T\cdot X_1 + ... + \phi_K(X)^T\cdot X_K
\end{equation}
with each $\phi_k(\cdot)$ being the $k$-th of $K$ output neurons of a standard feedforward Neural Network. Replacing the neural network by $K$ GP regression models, we arrive at a self-explaining Bayesian variant which was introduced by \citet{selfexpgps} under the term \textit{GPX}. In order to proceed, we now provide a brief recap on GP models and variational approximations in the following before exposing our main contributions.

\section{Gaussian Processes}
The building blocks of GPs, see \citet{rasmussengaussprocs}, are a prior distribution over functions, $p(f)$, and a likelihood $p(y|f)$. Using Bayes' law, we are interested in a posterior distribution $p(f|y)$ obtained as 

\begin{equation}\label{bayeslaw}
    p(f|y)=\frac{p(y|f)p(f)}{p(y)}.
\end{equation}

The prior distribution is a Gaussian Process, fully specified by $m(\cdot):\mathcal{X}\mapsto\mathbb{R}$, typically $m(x)=0$,  and covariance kernel function $k(\cdot,\cdot):\mathcal{X}\times\mathcal{X}\mapsto \mathbb{R}_0^+$:

\begin{equation}\label{gaussproc}
    p(f)=\mathcal{GP}(f|m(\cdot),k(\cdot,\cdot))
\end{equation}

We assume the input domain for $f$ to be a bounded subset of the real numbers, $\mathcal{X}\subset\mathbb{R}^K$. A common choice for $k(\cdot,\cdot)$ is the ARD-kernel

\begin{equation}\label{ardkern}
    k_{ARD}(x,x')=\theta\cdot exp(-0.5 (x-x')\Sigma(x-x')))
\end{equation}

where $\Sigma=diag(l_1^2,...,l_K^2)$ is a diagonal matrix with entries in $\mathbb{R}^+_0$ and $\theta>0$. For $K=1$, \eqref{ardkern} is equivalent to an SE-kernel. We denote by $K$ the positive semi-definite \textit{Gram-Matrix}, obtained as $K_{(ij)}=k(x_i,x_j)$, $x_i$ the $i$-th row of training input matrix $X_N$ containing $N$ observations in total, and write $K_{NN}$ for the Gram-Matrix over $X_N$.

Provided that $p(y|f)=\prod_{i=1}^N \mathcal{N}(y_i|f_i,\sigma^2)$, i.e. observations are i.i.d. univariate Gaussian conditioned on $f$, it is possible to directly calculate a corresponding posterior distribution for new inputs $X_*$ as

\begin{equation}\label{postpred}
    p(f_*|y)=\mathcal{MVN}(f_*|\tilde{\Lambda} y, K_{**}-\tilde{\Lambda} (K_{NN}+I\sigma^2) \tilde{\Lambda}^T)
\end{equation}

where $\tilde{\Lambda}=K_{*N}(K_{NN}+I\sigma^2)^{-1}$, $K_{*N,(ij)}=k(x^*_i,x_j)$, $K_{**,(ij)}=k(x^*_i,x^*_j)$; $I$ is the identity matrix with according dimension. 

In order to make GPs feasible for large datasets, the work of \cite{titsiassvgp, hensmanbig, hensmansvgpc} developed and refined Sparse Variational Gaussian Processes (SVGPs). SVGPs, introduce a set of $M$ so called inducing locations $Z_M\subset \mathcal{X}$ and corresponding inducing variables $f_M$. The resulting posterior distribution, $p(f,f_M|y)$, is then approximated through a variational distribution $q(f,f_M)=p(f|f_M)q(f_M)$ - usually $q(f_M)=\mathcal{N}(f_M|a,S),S=L L^T$ - by maximizing the \textit{evidence lower bound} (ELBO):

\begin{equation}\label{standardelbo}
    ELBO = \sum_{i=1}^n\mathbb{E}_{p(f|f_M)q(f_M)}\left[\log p(y_{(i)}|f_{(i)})\right]-KL(q(f_M)||p(f_M))
\end{equation}

where we obtain $p(f_M)$ by evaluating the GP prior distribution at inducing locations $Z_M$.
Following standard results for Gaussian random variables, it can also be shown that for marginal $q(f)$ evaluated at arbitrary $X_N$ and with $\Lambda=K_{NM}K_{MM}^{-1}$, we have

\begin{equation}\label{marginalvariationalfunction}
    q(f)=\mathcal{N}(f|\Lambda a,K_{NN}-\Lambda (K_{MM}-S)\Lambda^T)
\end{equation}

\section{Self-explaining variational posterior distributions}
Instead of formulating the prior GP model and subsequently deriving its variational approximation, we will proceed the opposite way by formulating the general structure of our variational distribution first. Using \eqref{senn} as a starting point for our model, an obvious adaption can be achieved by replacing the neural network with $K$ GPs, each modeling one corresponding varying coefficient. This is in direct relation to \citet{selfexpgps} who construct a self-explaining GP prior in this manner, coined \textit{GPX}. Such prior can be shown to yield a closed form posterior distribution of the same structure. Hence, part of our work can be seen as a variational extension to their method. As will be seen however, our method allows for extensions whose relation to the former is not as obvious. We will refer to our method as \textit{SEVGP} - \textbf{S}elf-\textbf{e}xplaining \textbf{v}ariational \textbf{G}aussian \textbf{P}rocess from now on.

The SEVGP is composed as follows - we have:

\begin{enumerate}
    \item $K$ independent sets of $M$ inducing variables at inducing locations $Z_M^{(k)}$, $\tilde{f}_M^{(k)},k\in\{1,...,K\}$, each corresponding to a separate GP
    \item $K$ independent GPs, $f_M^{(k)}$, with inducing locations as defined in 1.
    \item The actual target process, $f$, evaluated at arbitrary input matrix $X$ and constructed as
    
    \begin{equation}\label{sumconstruction}
        f(X)=\sum_{k=1}^{K}f_M^{(k)}(X)\odot X_k
    \end{equation}
    
    with $X_k$ the $k$-th column of $X$ and $\odot$ denoting element-wise multiplication.
\end{enumerate}

We can formally write the joint probability density via the conditional distributions

\begin{equation}\label{jointvariational}
    q(f,f_M^{(1)},...,f_M^{(K)},\tilde{f}^{(1)}_M,...,\tilde{f}^{(K)}_M) = q(f|f_M)\prod_{k=1}^{K}q(f^{(k)}_M|\tilde{f}^{(k)}_M)q(\tilde{f}^{(k)}_M)
\end{equation}

where we summarized $f_M=f_M^{(1)},...,f_M^{(K)},\,\,\tilde{f}_M=\tilde{f}_M^{(1)},...,\tilde{f}_M^{(K)}$ for convenience. Also we wrote $q(\cdot)$ instead of $p(\cdot)$ to stress that \eqref{jointvariational} is the target structure for the variational distribution for now.

Choosing $q(\tilde{f}^{(k)}_M)=\mathcal{N}(\tilde{f}^{(k)}_M|a^{(k)},S^{(k)})$ as in the standard SVGP model, we obtain in correspondence to \eqref{marginalvariationalfunction}:

\begin{equation}\label{marginalsummandproc}
        \begin{gathered}
        q(f_M^{(k)})=\int q(f_M^{(k)}|\tilde{f}_M^{(k)}) q(\tilde{f}_M^{(k)}) d\tilde{f}^{(k)}_M \\
        =\mathcal{N}(f_M^{(k)})|\Lambda^{(k)}a^{(k)},K^{(k)}_{NN}-\Lambda^{(k)}(K^{(k)}_{MM}-S^{(k)})\Lambda^{(k)T})
        \end{gathered}
\end{equation}

Defining $\mu^{(k)}=\Lambda^{(k)}a^{(k)}, \Sigma^{(k)}=K^{(k)}_{NN}-\Lambda^{(k)}(K^{(k)}_{MM}-S^{k})\Lambda^{(k)T}$, we can derive the marginal distribution of the target GP as 

\begin{equation}\label{variationalposterior}
    \begin{gathered}
        q(f)=\mathcal{N}(f|\sum_{k=1}^K \mu^{(k)}\odot X_k, \sum_{k=1}^K\Sigma^{(k)}\odot X_kX_k^T)\\
        =\mathcal{N}(f|\mu, \Sigma)
    \end{gathered}
\end{equation}

From the construction of our model it also follows that $q(f|f_M,\tilde{f}_M)=q(f|f_M)$, i.e. $f$ relates to $\tilde{f}_M$ only via $f_M$. We therefore conclude that

\begin{equation}\label{margproccondind}
    \begin{gathered}
    q(f|\tilde{f}_M)=\int q(f|f_M)\prod_{k=1}^{K}q(f^{(k)}_M|\tilde{f}^{(k)}_M)q(\tilde{f}^{(k)}_M) df_M \\
    = \mathcal{N}(f|\sum_{k=1}^{K} \Lambda^{(k)}\tilde{f}^{(k)}_M \odot X_k, \sum_{k=1}^{K} (K_{NN}^{(k)}-\Lambda^{(k)}K_{MM}^{(k)}\Lambda^{(k)T}) \odot X_kX_k^T)
    \end{gathered}
\end{equation}

where $df_M=df_M^{(1)}\cdots df_M^{(K)}$.

In addition, it is straightforward to see that under \eqref{margproccondind} we have

\begin{equation}\label{conditional_y}
    p(y|\tilde{f}_M)=\mathcal{N}(y|\sum_{k=1}^{K} \Lambda^{(k)}\tilde{f}^{(k)}_M \odot X_k, \sum_{k=1}^{K} (K_{NN}^{(k)}-\Lambda^{(k)}K_{MM}^{(k)}\Lambda^{(k)T}) \odot X_kX_k^T + I\sigma^2)
\end{equation}

Equations \eqref{marginalsummandproc}, \eqref{variationalposterior} and \eqref{margproccondind} now allow to construct different prior distributions and hence express different prior beliefs. We remind the reader that the paramount goal of all three approaches is expression of meaningful prior belief on the one hand and transparency of the result on the other hand. While the latter has been exemplified in this section, the former is achieved by three different interpretations of the proposed variational posterior. We provide a tabular overview of all three variants in \textbf{Appendix A}.

\subsection{As a variational extension for GPX}
As stated above, the most obvious interpretation is as a sparse variational approximation of a GPX model. There are two alternative ways to implement this variant:

\begin{enumerate}
    \item Have both prior and variational process structured as in \eqref{variationalposterior} and perform variational inference for SVGPs as usual. 
    \item Treat each varying coefficient GP separately, i.e. each GP $f_M^{(1)},...,f_M^{(K)}$ has its own set of inducing points and variables 
\end{enumerate}

As the former case would be trivial and not help us incorporating any meaningful prior knowledge about the coefficients, we focus on the latter. This case can be easily derived by constructing the prior conditioned on its realizations at $Z_M=\{Z_M^{(1)},...,Z_M^{(K)}\}$, $p(f|f_M)$, as in \eqref{margproccondind}, i.e. 

\begin{equation}\label{priorselfexp}
    p(f|\tilde{f}_M)=\mathcal{N}(f|\sum_{k=1}^{K} \Lambda^{(k)}\tilde{f}^{(k)}_M \odot X_k, \sum_{k=1}^{K} (K_{NN}^{(k)}-\Lambda^{(k)}K_{MM}^{(k)}\Lambda^{(k)T}) \odot X_kX_k^T).
\end{equation}

Adding now $p(\tilde{f}_M)$, the joint prior distribution over $f,\tilde{f}_M$ is

\begin{equation}\label{gpxvarprior}
    p(f,\tilde{f}{}_M)=p(f|\tilde{f}_M)p(\tilde{f}_M)=p(f|\tilde{f}_M)\prod_{k=1}^{K}p(\tilde{f}_M^{(k)}).
\end{equation}

The variational distribution is slightly modified to match the standard SVGP structure per process $k$:

\begin{equation}\label{gpxvarvariational}
    q(f,\tilde{f}_M)=p(f|\tilde{f}_M)q(\tilde{f}_M)=p(f|\tilde{f}_M)\prod_{k=1}^{K}\mathcal{N}(\tilde{f}_M^{(k)}|a^{(k)},S^{(k)}).
\end{equation}

Using \eqref{gpxvarprior} and \eqref{gpxvarvariational}, we conduct the variational posterior approximation via

\begin{equation}\label{kldiv41}
    KL_{4.1}=KL(q(f,\tilde{f}_M)||p(f,\tilde{f}_M|y))
\end{equation}

Equation \eqref{kldiv41} then results in the following ELBO - the derivation of this result and all subsequent ones can be found in the appendix:

\begin{equation}\label{elbo41}
    ELBO_{4.1}=\sum_{i=1}^N\log\mathcal{N}(y_i|\mu_i,\sigma^2)-\frac{1}{2\sigma^2}tr(\Sigma)-\sum_{k=1}^K KL(q(f_M^{(k)})||p(f_M^{(k)}))
\end{equation}

It should be clear that \eqref{elbo41} also implies that we can use batch sampling methods in order to apply our method to large datasets. 

This rather obvious SVGP extension to GPX allows to incorporate prior knowledge about each of the varying coefficients or respectively each feature's contribution individually.

\subsection{As a variational approximation for an arbitrary GP}
In order to allow for general functional prior knowledge, we now use \eqref{variationalposterior} directly and approximate an arbitrary GP posterior $p(f|y)$. As a crucial distinction to plain SVGPs, we allow the covariance functions of $q(f)$ and $p(f|y)$ to be different while keeping the structure of $q(f)$ self-explaining as before.

Hence, the KL-objective for variational inference in this case becomes

\begin{equation}\label{kldiv42}
    KL_{4.2}=KL(q(f)||p(f|y)).
\end{equation}

Since \eqref{kldiv42} denotes a KL-divergence between two stochastic processes, we cannot proceed as in the finite dimensional case as noted in \citet{funcvarinf}. As a result, we cannot obtain a usual ELBO either but derive a \textit{functional evidence lower bound} (fELBO) instead:

\begin{equation}\label{felbo42}
    fELBO_{4.2}=\sum_{i=1}^N\log\mathcal{N}(y_i|\mu_i,\sigma^2)-\frac{1}{2\sigma^2}tr(\Sigma_N)- KL(q(f^{\{N,A\}})||p(f^{\{N,A\}}))
\end{equation}

with $p(f^{\{N,A\}}),q(f^{\{N,A\}})$ the finite dimensional evaluation of $p(f),q(f)$ over the union set consisting of training observations $N$ and so called augmentation points $A$ sampled uniformly from $\mathcal{X}$. Also, $\Sigma_N$ denotes the evaluation of the finite dimensional covariance matrix of $q(f)$ evaluated at $N$

As long as the component kernel functions of $q(f)$ are flexible enough, it is possible to approximate a large variety of prior functions with this approach. To stress the difference of this approach to standard SVGPs, the resulting variational posterior allows for case-based explanations for each instance predicted.

\subsection{As a variational approximation for a GP with additional priors over coefficients}
This last variant can be interpreted as a combination of the other two approaches. In a corresponding use-case scenario we might have prior knowledge available on both the functional form over all input variables and, additionally, specific prior knowledge about the individual contribution of certain features. To embed this idea into our framework, we make the following adjustments to the prior and variational distributions from \eqref{gpxvarprior} and \eqref{gpxvarvariational}:

The modified prior distribution is now structured as

\begin{equation}\label{priorstruct43}
    p(f,\tilde{f}_M)=p(f)p(\tilde{f}_M)=p(f)\prod_{k=1}^{K}p(\tilde{f}_M^{(k)}),
\end{equation}

the modified variational distribution as

\begin{equation}\label{variationalstruct43}
    q(f,\tilde{f}_M)=q(f|\tilde{f}_M)q(\tilde{f}_M)=q(f|\tilde{f}_M)\prod_{k=1}^{K}q(\tilde{f}_M^{(k)}),
\end{equation}

As the resulting ELBO formula facilitates the explanation of our reasoning behind \eqref{priorstruct43} and \eqref{variationalstruct43}, we first state the resulting variational objective. For the KL-divergence we have

\begin{equation}\label{kldiv43}
    KL_{4.3}=KL(q(f,\tilde{f}_M)||p(f,\tilde{f}_M|y))
\end{equation}

As it turns out, we require a functional lower bound in this case as well:

\begin{equation}\label{felbo43}
    \begin{gathered}
        fELBO_{4.3}=\sum_{i=1}^N\log\mathcal{N}(y_i|\mu_i,\sigma^2)-\frac{1}{2\sigma^2}tr(\Sigma_N)\\
        -KL(q(f^{\{N,A\}})||p(f^{\{N,A\}}))-\frac{1}{2}tr(C_{\{N,A\}}^{-1}\Sigma_{\{N,A\}})-\sum_{k=1}^K KL(q(\tilde{f}_M^{(k)})||p(\tilde{f}_M^{(k)}))
    \end{gathered}
\end{equation}

$C_{\{N,A\}}$ denotes the evaluation of the finite dimensional covariance matrix of $p(f)$ at ${\{N,A\}}$.  

Assuming independence between $\tilde{f}_M$ and $f$ in the prior distribution while keeping them dependent in the variational approximation requires some explanation. Presuming that the true data-generating function is unlikely to follow an additive structure as postulated in \eqref{priorselfexp}, there is no reason to split $f$ into additive components in the general case. Hence, we treat $p(f)$ as an independent process, irrespective of any supplemental $f_M$ and corresponding inducing variables. In correspondence to the construction discussed at the beginning, this implies

\begin{equation}\label{margproccondind43}
    \begin{gathered}
        p(f,\tilde{f}_M)=\int p(f|f_M)p(f^{(k)}_M|\tilde{f}^{(k)}_M)p(\tilde{f}^{(k)}_M) df_M\\
        =\int p(f)\prod_{k=1}^{K}p(f^{(k)}_M|\tilde{f}^{(k)}_M)p(\tilde{f}^{(k)}_M) df_M=p(f)\prod_{k=1}^{K}p(\tilde{f}^{(k)}_M)
    \end{gathered}
\end{equation}

in accordance with \eqref{priorstruct43}. The discussed prior distribution is hence rather a convenience in order to influence both $q(f)$ as a whole and each $q(f|\tilde{f}_M)$ individually while at the same time conserving the transparent structure of the variational distribution.

\section{Experiments}
In order to empirically evaluate our method, we first conducted an experiment on posterior soundness to validate for a simple example the properties we discussed in section 4. The choice of coefficient priors $p(f_m^{(k)})$ and full process prior $p(f)$ should be reflected accordingly in the corresponding posteriors of each variant. As a second experiment, we compared the results of SEVGP with SENN on standard regression datasets. 

More implementation details can be found in \textbf{Appendix C}.

\subsection{Posterior soundness}

We sampled input data $X$ uniformly from $[-2,2]$ and generated the target feature as $y=0.25x^2+\epsilon,\,\epsilon\sim\mathcal{N}(0,0.25)$. For the varying coefficient prior, ($p(f^{(k)}_m)$, variants $4.1$ and $4.3$) and variational ($q(f^{(k)}_m)$, all three variants) processes a $GP$ with zero mean. As a kernel function a, we used a summation kernel of a constant kernel $k_{const}(x,x')=1$ and an SE-kernel $k_{SE}=a\cdot exp\left(-\frac{(x-x')^2}{l^2}\right)$ and $l$ the only trainable parameter. While the former corresponds to the prior belief of the varying coefficient being constant, the latter exemplifies potential deviation from constancy. The trade-off between both assumptions can be steered via $a$. For our experiments, we set $a=0.5$. 

\begin{figure}
\centering
\begin{subfigure}{.5\textwidth}
  \centering
  \includegraphics[width=1.\linewidth]{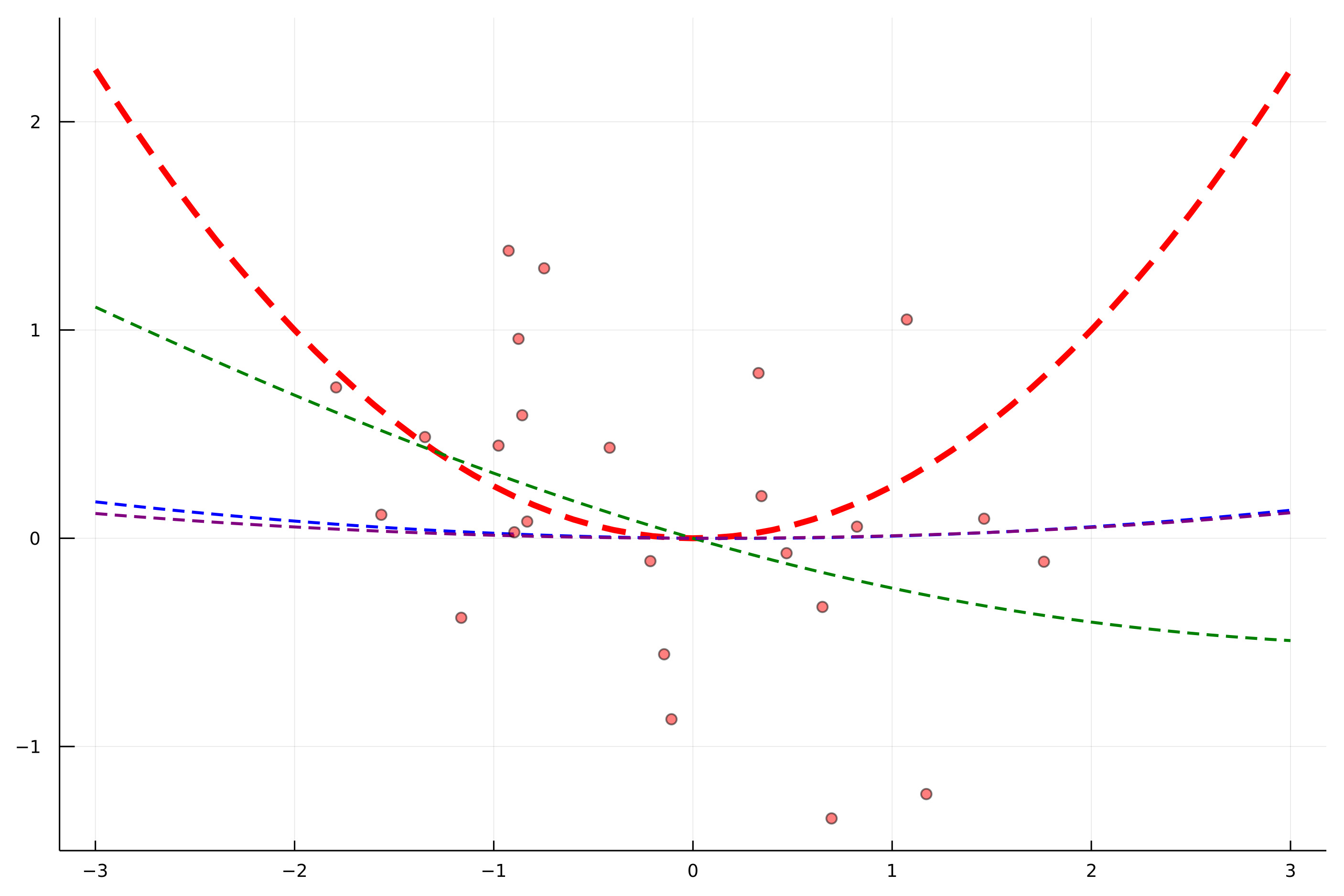}
  \label{fig:meanvalue:sub1}
\end{subfigure}%
\begin{subfigure}{.5\textwidth}
  \centering
  \includegraphics[width=1.\linewidth]{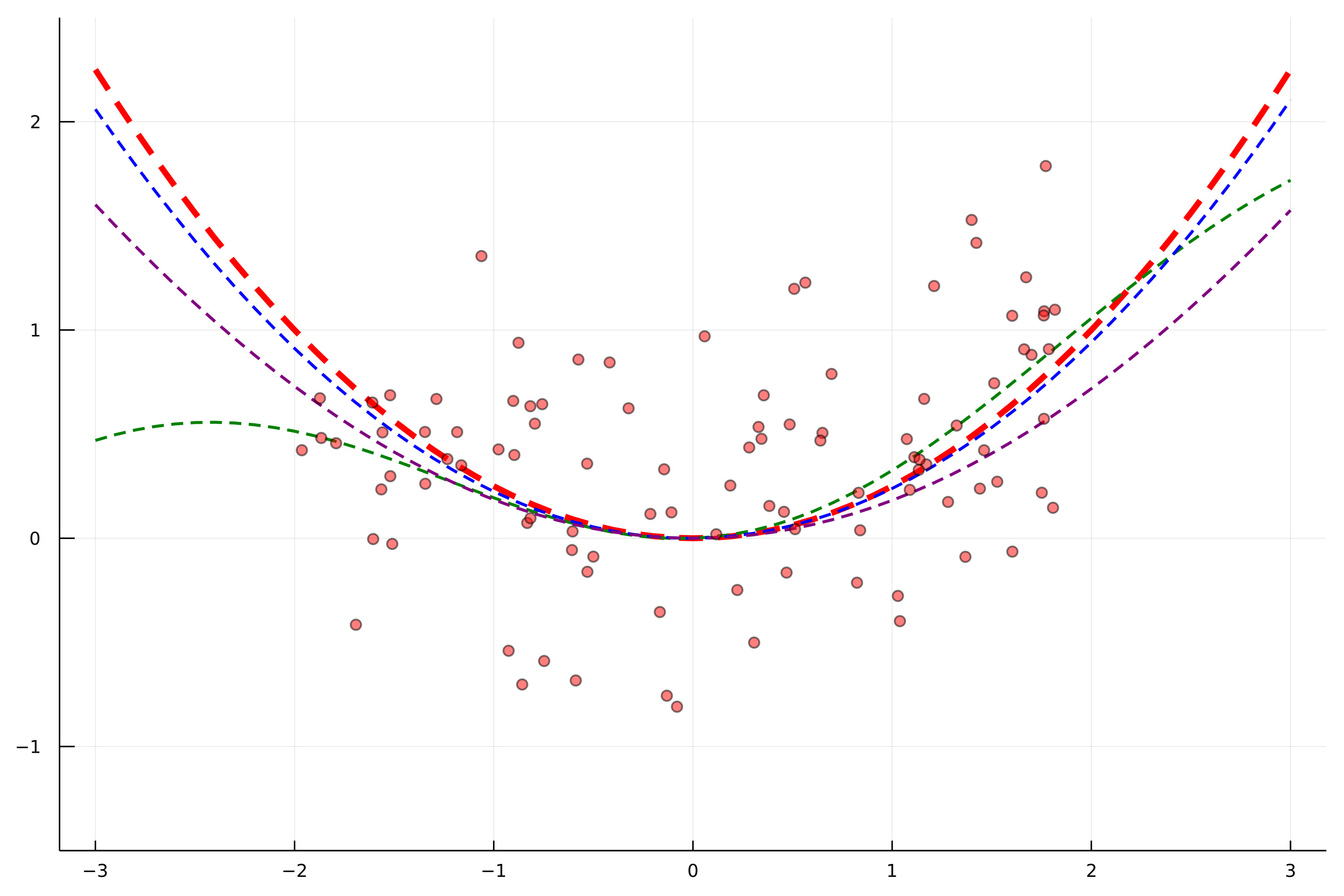}
  \label{fig:meanvalue:sub2}
\end{subfigure}
\caption{Variational posterior predictive mean functions for variants $4.1$ (green), $4.2$ (blue), $4.3$ (purple) for samples (red dots) of size $N_1=25$ (left) and $N_2=100$ (right) with true mean function (red) $y=0.25x^2$.}
\label{fig:meanvalue}
\end{figure}

For variants $4.2,4.3$, we posed a second-order polynomial GP prior over the full process, $p(f)=\mathcal{GP}(f|0,k_{poly}(\cdot,\cdot)),\,k_{poly}(x,x')=(x \cdot x')^2$ - i.e. our prior belief matches the data generating process. As can be seen in \figref{fig:meanvalue}, the posterior means for these variants approach the true mean function more quickly. In the larger sample case, the posterior means under the polynomial prior better fit the true mean outside the range of observed training examples. As for the difference between variants $4.2$ and $4.3$, we observed that the additional prior $p(f_m)$ in $4.3$ yielded a regularizing effect on the posterior distribution.

\subsection{SENN-comparison}

In order to evaluate the practical applicability of our approach, we compared it against a SENN as in \citet{selfexplainnets} on four UCI datasets\footnote{Boston Housing, Concrete Slump, Red Whine, White Whine, see \citet{Dua:2019}}. It should be stressed that our aim was not to show that our method provides better results in general but rather that its performance is comparable to SENN. Also, we only included variant $4.1$ in this evaluation given that we had no sensible prior knowledge about $p(f)$ available for these datasets.

We ran 10-fold cross-validation per dataset and evaluated both mean-squared error (MSE) and coefficient stability, i.e. explanation coherence for neighboring datapoints. To evaluate both measures for the GP models, we used the posterior mean function. In regards to stability, we averaged over all training examples the following stability measure per instance $x_i$:

\begin{equation}\label{stability}
    L(x_i)=\max_{x_j\in\hat{B}_{10}(x_i)}\frac{||F_m(x_i)-F_m(x_j)||_2}{||x_i-x_j||_2}
\end{equation}

\begin{table}
    \caption{MSE and coefficient stability for SENN and SEVGP (variant 4.1) posterior mean; average and standard deviation over 10-fold cross validation}
    \label{table:evaluation}
    \centering
    \begin{tabular}{|l|l|l|l|l|}
    \hline
    \multirow{4}{*} &
      \multicolumn{2}{c|}{\textbf{{\small SENN}}} &
      \multicolumn{2}{c|}{\textbf{{\small SEVGP (4.1)}} (ours)} \\
    & {\footnotesize MSE} & {\footnotesize Stability} & {\footnotesize MSE} & {\footnotesize Stability} \\
    \hline
    { Boston} & {$0.1325\pm 0.0637$} & {$0.2785\pm0.0208$} & {$0.1655\pm 0.0724$} & {$0.2150\pm0.0178$} \\
    \hline
    { Concrete} &  {$0.0259\pm 0.0215$} & {$0.2780\pm0.0364$} & {$0.0196\pm 0.0109$} & {$0.2118\pm0.0222$} \\
    \hline
    { Wine red} & {$0.6871\pm 0.0928$} & {$0.2538\pm0.0275$} & {$0.6362\pm 0.0531$} & {$0.0697\pm0.0138$} \\
    \hline
    { Wine white} &  {$0.7360\pm0.1959$} & {$0.3153\pm0.0615$} & {$0.6858\pm0.0383$} & {$0.0836\pm0.0135$} \\
    \hline
    \end{tabular}
\end{table}
    
with $F_m(x_i)$ the stacked vector of coefficients at $x_i$ derived either from the GP posterior mean or the SENN output neurons. $\hat{B}_{m}(x_i)$ denotes the set of the $m$ training instances with $x_j\neq x_i$ closest to $x_i$\footnote{\citet{selfexpgps} propose to use all training instances in the $\epsilon$-neighorhood of $x_i$ for their stability measure. For our datasets this lead to issues due to outliers, hence we resorted to this variant.}. As a kernel function for the SEVGP coefficient priors, we used the same kernel as in \textbf{5.1} but replaced the SE-kernel by an ARD-kernel \eqref{ardkern} and set $a=2$ to allow for more variability in the coefficients.

Table \ref{table:evaluation} shows that SEVGP achieves comparable performance to SENN in terms of MSE and, particularly, stability.

\section{Related work}
The results of \citet{selfexplainnets,selfexpgps,distributedvarcoeff} directly inspired our approach from an explainability and transparency point of view. The overarching, general theme of our present work however revolves around the question of how to make prior knowledge available to complex Machine Learning models. \citet{virtualexamples,valuepriorknowledge,informedml,quantitativedomain} all discuss the potentially beneficial role of expert and domain knowledge in Machine Learning, yet either mention Bayesian methods only briefly or not at all. Nevertheless, Bayesian non-parametrics have already been applied successfully in countless classical statistical modeling problems with an emphasis on incorporating prior knowledge - see \citet{gelman2013bayesian} for a variety of examples. 

Recent work on functional variational inference as discussed particularly in \citet{funcvarinf, understandingvarinf} could be a fruitful step towards a synthesis of meaningful prior models and modern Machine Learning architectures. On another note, \citet{generalizedinference} re-interpret variational inference as a mere optimization problem where the KL-divergences in an evidence lower bound are merely seen as regularization terms. This view is apparently related to our derivations in 4.3 where we explicitly dragged along the inducing variables in order to influence the posterior distribution based on prior knowledge about the coefficients and the overall function simultaneously.

\section{Limitations and discussion}
The main challenge we currently see for our method is its scalability to more complex problems. In particular, image classification tasks could potentially benefit from more expressive prior distributions, especially when training datasets are difficult to populate or diversify. This also distinguishes the proposed model from its neural counterpart in \citet{selfexplainnets} which can be scaled quite efficiently. Nevertheless, the sparse variational setup will likely allow for further improvements in terms of scalability. With this in mind, we are confident that our method can be extended to computer vision tasks in the future.

On a broader scale, the general possibilities to interweave Bayesian methods with transparent machine learning might be far from being exhausted with our contribution. As the field of interactive and human-in-the-loop machine learning - see  \citet{fails2003interactive,whendoweneedhumanloop} - is getting more traction, such methods could be used for transparent human-machine feedback loops as proposed for example by \citet{explainableinteractive}.

\section{Broader impact}
The potential societal impact that we hope to contribute to with this work is two-fold: First, the self-explaining structure of our approach allows for transparency about which features or variables contributed to a given outcome. This allows to check whether a prediction is driven by any form of model bias and unfairness at runtime. On the flipside, we haven't worked out any global guarantees for fairness and unbiasedness in this work. 

The second possible impact of our method is the ability to reduce the chance of unfairness and biasedness from the start through the choice of prior distributions on the varying coefficients. By using for example a sufficiently positive prior mean function for a coefficient, we can guide the corresponding posterior coefficient to take on positive values with high probability as well, resulting in a positive effect for a given feature. However, there are again no guarantees and the resulting predictions would still need to be validated.

On the negative side of impact, we see no actively harmful potential of our method. However, a  user might falsely presume these properties to be present. A subsequent negative impact would thus be highly dependent on the specific use-case at hand and could occur with any method whenever such methodological misunderstanding prevails. To prevent such issues, any user of our method should be made aware of these shortcomings.

\medskip
\bibliographystyle{plainnat}
\bibliography{references}

\newpage
\appendix

\section{Overview over prior structures}
\begin{center}
\def\arraystretch{1.25}
\begin{tabular}{ | M{4em} | M{16em}| M{12em} | } 
\hline
\textbf{Variant} & \textbf{Lower Bound} & \textbf{Implied prior knowledge} \\ 
\hline
4.1 & $ELBO_{4.1}=\sum_{i=1}^N\log\mathcal{N}(y_i|\mu_i,\sigma^2)-\frac{1}{2\sigma^2}tr(\Sigma)-\sum_{k=1}^K KL(q(f_M^{(k)})||p(f_M^{(k)}))$ & Knowledge about individual variables/coefficients \\ 
\hline
4.2 & $fELBO_{4.2}=\sum_{i=1}^N\log\mathcal{N}(y_i|\mu_i,\sigma^2)-\frac{1}{2\sigma^2}tr(\Sigma_N)- KL(q(f^{\{N,A\}})||p(f^{\{N,A\}}))$ & Knowledge about general functional relation\\ 
\hline
4.3 & $fELBO_{4.3}=\sum_{i=1}^N\log\mathcal{N}(y_i|\mu_i,\sigma^2)-\frac{1}{2\sigma^2}tr(\Sigma_N)KL(q(f^{\{N,A\}})||p(f^{\{N,A\}}))-\frac{1}{2}tr(C_{\{N,A\}}^{-1}\Sigma_{\{N,A\}})-\sum_{k=1}^K KL(q(\tilde{f}_M^{(k)})||p(\tilde{f}_M^{(k)}))$ & Knowledge about general functional relation and individual variables/coefficients\\ 
\hline
\end{tabular}
\end{center}

\FloatBarrier

\section{Derivations}
We first state the following well-known result for the expected sum of log-likelihoods of i.i.d univariate Gaussians with respect to a common mean $M$ vector with multivariate Gaussian distribution:
\begin{equation}\label{wellknownresult}
    \begin{gathered}
        \mathbb{E}_{M\sim\mathcal{N}(\mu,\Lambda)}\left[\sum_{i=1}^N \log \mathcal{N}(y_i|M_i,\sigma^2)\right]\\
        =\sum_{i=1}^N \log\mathcal{N}(y_i|\mu_i,\sigma^2)-\frac{1}{2\sigma^2}tr(\Lambda)
    \end{gathered}
\end{equation}

In order to approximate a stochastic process posterior $p(f|y)$ through variational process $q(f)$, the following theorem from \citet{funcvarinf} allows to construct a lower bound via the finite dimensional measurement set $D$:

\begin{theorem}\label{theorem2funcvar}
    If $D$ contains all training inputs $N$, then 
    
    $$\log p(y)\geq KL(q(f^D)|p(f^D|y))$$
\end{theorem}

We refer to the above work for the proof of this theorem.

\subsection{Derivation of $ELBO_{4.1}$}

$$KL_{4.1}=\int q(f,\tilde{f}_M)\log\frac{q(f,\tilde{f}_M)}{p(f,\tilde{f}_M|y)} df d\tilde{f}_M$$

$$=\int q(f,\tilde{f}_M)\log\frac{q(f,\tilde{f}_M)p(y)}{p(y|f,\tilde{f}_M)p(f,\tilde{f}_M)} df d\tilde{f}_M$$

$$=\int q(f,\tilde{f}_M)\log\frac{p(f|\tilde{f}_M)q(\tilde{f}_M)p(y)}{p(y|f,\tilde{f}_M)p(f|\tilde{f}_M)p(\tilde{f}_M)} df d\tilde{f}_M$$

$$=\int q(f,\tilde{f}_M)\log\frac{\cancel{p(f|\tilde{f}_M)}q(\tilde{f}_M)p(y)}{{p(y|f,\tilde{f}_M)}\cancel{p(f|\tilde{f}_M)}p(\tilde{f}_M)} df d\tilde{f}_M$$

$$=\int p(f|\tilde{f}_M)q(\tilde{f}_M)\log\frac{q(\tilde{f}_M)p(y)}{{p(y|f,\tilde{f}_M)}p(\tilde{f}_M)} df d\tilde{f}_M$$

$$=\int p(f|\tilde{f}_M)q(\tilde{f}_M)\log\frac{q(\tilde{f}_M)}{p(\tilde{f}_M)} df d\tilde{f}_M-\int p(f|\tilde{f}_M)q(\tilde{f}_M)\log p(y|f,\tilde{f}_M)df d\tilde{f}_M+\log p(y)$$

$$=\int p(f|\tilde{f}_M)\prod_{k=1}^Kq(\tilde{f}^{(k)}_M)\log\frac{\prod_{k=1}^Kq(\tilde{f}^{(k)}_M)}{\prod_{k=1}^Kp(\tilde{f}^{(k)}_M)} df d\tilde{f}_M-\int p(f|\tilde{f}_M)q(\tilde{f}_M)\log p(y|f,\tilde{f}_M)df d\tilde{f}_M+\log(p)$$

$$=\sum_{k=1}^K KL(q(f_M^{(k)})||p(f_M^{(k)}))-\int p(f|\tilde{f}_M)q(\tilde{f}_M)\log p(y|f,\tilde{f}_M)df d\tilde{f}_M+\log(p)$$

$$\stackrel{\eqref{conditional_y}}{=}\sum_{k=1}^K KL(q(f_M^{(k)})||p(f_M^{(k)}))-\int p(f|\tilde{f}_M)q(\tilde{f}_M)\log p(y|f,\tilde{f}_M)df d\tilde{f}_M+\log(p)$$

$$=\sum_{k=1}^K KL(q(f_M^{(k)})||p(f_M^{(k)}))-\mathbb{E}_{q(\tilde{f}_M)}\left[\sum_{i=1}^N \log p(y_i|\tilde{f}_M)\right]+\log(p)$$

$$\stackrel{\eqref{variationalposterior},\eqref{wellknownresult}}{=}\sum_{k=1}^K KL(q(f_M^{(k)})||p(f_M^{(k)}))-\sum_{i=1}^N \log\mathcal{N}(y_i|\mu_i,\sigma^2)+\frac{1}{2\sigma^2}tr(\Sigma)+\log(p)$$

$$\Rightarrow ELBO_{4.1}=\sum_{i=1}^N\log\mathcal{N}(y_i|\mu_i,\sigma^2)-\frac{1}{2\sigma^2}tr(\Sigma)-\sum_{k=1}^K KL(q(f_M^{(k)})||p(f_M^{(k)}))$$

\subsection{Derivation of $fELBO_{4.2}$}

$fELBO_{4.2}$ is directly obtained from \textbf{Theorem 1} by plugging in our GP definitions for $p(f),q(f)$ and by applying \eqref{wellknownresult}.

\subsection{Derivation of $fELBO_{4.3}$}

We follow the origina proof for \textbf{Theorem 1} and show that the proposed $fELBO_{4.3}$ is indeed a lower bound for $\log p(y)$ under the stated assumptions. We first state the following result which we will prove later on:

$$\mathbb{E}_{q(\tilde{f}_M)}[KL(q(f^{\{N,A\}}|\tilde{f}_M)||p(f^{\{N,A\}}))]=KL(q(f^{\{N,A\}})||p(f^{\{N,A\}}))+\frac{1}{2}tr(C_{\{N,A\}}^{-1}\Sigma_{\{N,A\}})$$

This allows us to write the corresponding $fELBO$ as

$$fELBO_{4.3}$$
$$=\sum_{i=1}^N\log\mathcal{N}(y_i|\mu_i,\sigma^2)-\frac{1}{2\sigma^2}tr(\Sigma)-\mathbb{E}_{q(\tilde{f}_M)}[KL(q(f^{\{N,A\}}|\tilde{f}_M)||p(f^{\{N,A\}}))]-\sum_{k=1}^K KL(q(\tilde{f}_M^{(k)})||p(\tilde{f}_M^{(k)}))$$

$$=\mathbb{E}_{q(f^D,\tilde{f}_M)}[\log p(y|f^D)-\log q(f^D|\tilde{f}_M)+\log p(f^D)-\log q(\tilde{f}_M)+\log p(\tilde{f}_M)]$$

where we summarized $D=\{N,A\}$ and used the fact that for observations $p(y|f)=p(y|f^N)$ the augmentation points $A$ are irrelevant, hence $p(y|f)=p(y|f^{\{N,A\}})$ holds.

$$=\log p(y)-\mathbb{E}_{q(f^D,\tilde{f}_M)}\left[\log\frac{q(f^D|\tilde{f}_M)q(\tilde{f}_M)p(y)}{p(y|f^D)p(f^D)p(\tilde{f}_M)}\right]$$

$$=\log p(y)-\mathbb{E}_{q(f^D,\tilde{f}_M)}\left[\log\frac{q(f^D,\tilde{f}_M)}{p(f^D,\tilde{f}_M|y)}\right]$$

where used the fact that $p(\tilde{f}_M)\indep p(f^D)$ by construction \eqref{priorstruct43}, hence $p(\tilde{f}_M)\indep p(y)$ and therefore $p(f^D,\tilde{f}_M|y)=p(f^D|y)p(\tilde{f}_M)$.

$$=\log p(y)-KL(q(f^D,\tilde{f}_M)||p(f^D,\tilde{f}_M|y))$$

$$\Rightarrow \log p(y)\geq fELBO_{4.3}$$
\linebreak

We next derive the closed form expression for $\mathbb{E}_{q(\tilde{f}_M)}[KL(q(f^{\{N,A\}}|\tilde{f}_M)||p(f^{\{N,A\}}))]$:

$$\mathbb{E}_{q(\tilde{f}_M)}[KL(q(f^{\{N,A\}}|\tilde{f}_M)||p(f^{\{N,A\}}))]$$

$$=\mathbb{E}_{q(\tilde{f}_M)}[KL(q(f^D|\tilde{f}_M)||p(f^D))]$$

\begin{equation}\label{tobecontinued}
    \mathbb{E}_{q(\tilde{f}_M)}\bigg\{\frac{1}{2}\left[\frac{\log|C_D|}{\log |\Sigma_{D}|}-tr(C_D^{-1}\Sigma_D)+(m_D-\mu_{|\tilde{f}_M})^TC_D^{-1}(m_D-\mu_{|\tilde{f}_M})\right]\bigg\}
\end{equation}

with $m_D, C_D$ the mean and kernel functions of $p(f)$ evaluated at $D$,

$$\mu_{|\tilde{f}_M}=\sum_{k=1}^{K} \Lambda_D^{(k)}\tilde{f}^{(k)}_M \odot X_k$$

the mean vector corresponding to $q(f|\tilde{f}_M)$ evaluated at $D$,

$$\Sigma_D=\sum_{k=1}^{K} (K_{D}^{(k)}-\Lambda^{(k)}_D K_{MM}^{(k)}\Lambda^{(k)T}_D) \odot X_kX_k^T$$

the kernel gram matrix corresponding to $q(f|\tilde{f}_M)$ evaluated at $D$. Continuing from \eqref{tobecontinued} we have:

$$=\frac{1}{2}\bigg[\frac{\log|C_D|}{\log |\Sigma_{X}|}-tr(C_D^{-1}\Sigma_D)\bigg]+\frac{1}{2}\mathbb{E}_{q(\tilde{f}_M)}\bigg[m_D^TC_D^{-1} m_D -2m_D^TC_D^{-1}\mu_{|\tilde{f}_M}+\mu_{|\tilde{f}_M}^TC_D^{-1}\mu_{|\tilde{f}_M}\bigg]$$

$$=\frac{1}{2}\bigg[\frac{\log|C_D|}{\log |\Sigma_{X}|}-tr(C_D^{-1}\Sigma_D)+m_D^TC_D^{-1} m_D-2m_D^TC_D^{-1}\mu\bigg]+\frac{1}{2}\mathbb{E}_{q(\tilde{f}_M)}\bigg[ \mu_{|\tilde{f}_M}^TC_D^{-1}\mu_{|\tilde{f}_M}\bigg]$$

$$=\frac{1}{2}\bigg[\frac{\log|C_D|}{\log |\Sigma_{X}|}-tr(C_D^{-1}\Sigma_D)+m_D^TC_D^{-1} m_D-2m_D^TC_D^{-1}\mu+\mu^TC_D^{-1}\mu\bigg]+\frac{1}{2}tr(C_D^{-1}\Sigma_D)$$

$$=KL(q(f^{\{N,A\}})||p(f^{\{N,A\}}))+\frac{1}{2}tr(C_{\{N,A\}}^{-1}\Sigma_{\{N,A\}})$$

\newpage

\section{Implementation details}
All experiments were performed on a MacBook Pro (2018), 2,2 GHz 6-Core Intel Core i7, 16 GB 2400 MHz DDR4 using primarily Julia 1.6 (see \citet{bezanson2017julia}) and Jupyter Lab (see \citet{kluyver2016jupyter}). Our library of choice for automatic differentiation and gradient based optimization is Zygote.jl as described in \citet{Zygote.jl-2018}. The exact version of each library can be found in the supplemental files. All packages are licensed as open-source.

\subsection{Implementation details 5.1}

We sampled from a joint-distribution $p(x,y)$ obtained as follows:

\begin{equation}
    \begin{gathered}
        p(x,y)=p(y|x)p(x)\\
        p(x)=\mathcal{U}(x|-2,2)\\
        p(y|x)=\mathcal{N}(y|0.25x^2,0.25)
    \end{gathered}
\end{equation}

In order to compare behavior for different sample sizes, we created two sets of $i.i.d$ samples from this distribution, the first having $N_1=25$ samples, the second having $N_2=100$ samples. For the sake of comparability of our results, we used the same structure for the variational posterior regardless of the variant we examined. In particular, we used a $GP(0,k_{const-se}(\cdot,\cdot))$ prior for $p(f_M)$ with

\begin{equation}\label{constsekernel}
    k_{const-se}(x,x')=1 + 0.5\cdot exp\left(-\frac{(x-x')^2}{l^2}\right)
\end{equation}

with $l^2$ the only trainable parameter. 

For variants $4.2,4.3$ we used a $GP(0,k_{poly}(\cdot,\cdot))$ prior for $p(f)$ with polynomial kernel

\begin{equation}
    k_{poly}(x,x')=(x\cdot x'^T)^2. 
\end{equation}

The number of inducing points per process $k$ was $|Z_m|=4$. Also we slightly adjusted the functional KL divergence $KL(q(f^D)||p(f^D))$ by adding a regularization parameter $\lambda$ as proposed by \citet{funcvarinf}. For $fELBO_{4.2}$, this results in 

\begin{equation}
    fELBO_{4.2}=\mathbb{E}_{q(f)}[p(y|f)]-\lambda\cdot KL(q(f^D)||p(f^D))
\end{equation}

and accordingly for $fEBLO_{4.3}$. In our experiments we set $\lambda=\frac{1}{N}$ which results in an improper lower bound. However, the posterior processes turned out to be much more stable with this setup.

\subsection{Implementation details 5.2}

For the SENN model, we used a 1-hidden layer feedforward neural network with 20 neurons in the hidden-layer and $SELU$ activation function. We used an ADAM optimizer for the SENN optimization.

For the SEVGP model, we used the same kernel as in \textbf{5.1} but used an ARD-kernel instead of the SE-kernel to account for multidimensionality in the datasets and increased the variance factor to $a=2$:

\begin{equation}\label{constsekernel}
    k_{const-se}(x,x')=1 + 2\cdot exp(-0.5 (x-x')\Sigma(x-x')))
\end{equation}

with $\Sigma$ a diagonal matrix with positive entries. We used an RMSProp optimizer for the SEVGP optimization. Given the size of some of the datasets we applied MC-sampling to estimate the mean gradient for the lower bound using a sample size of $100$. To ensure comparability, this was done for both SEVGP and SENN

The number of inducing points per process $k$ was $|Z_m|=3$.


\end{document}